\documentclass[12pt]{article}

\usepackage[utf8]{inputenc} 
\usepackage{xcolor}
\usepackage{amsmath}
\usepackage{amssymb}
\usepackage{amsfonts}
\usepackage{amsthm}
\usepackage{graphicx}
\usepackage[]{caption}
\usepackage{anyfontsize}
\usepackage[]{subcaption}
\usepackage[ruled]{algorithm2e}
\usepackage{booktabs}
\usepackage{multirow}
\usepackage{wrapfig}
\usepackage{marvosym}

\newcommand{\eps}{\varepsilon}
\newcommand{\dd}{\mathrm{d}}
\newcommand{\RR}{\mathbb{R}}

\newcommand{\EE}{\mathbb{E}}
\newcommand{\Var}{\mathrm{Var}}
\newcommand{\DD}{\mathcal{D}}
\newcommand{\Ncal}{\mathcal{N}}
\newcommand{\ssigma}{\boldsymbol{\sigma}}
\newcommand{\zzeta}{\boldsymbol{\zeta}}
\newcommand{\KL}{D_{\mathrm{KL}}}
\newcommand{\Lcal}{\mathcal{L}}

\DeclareMathOperator*{\argmin}{\mathrm{arg}\min}

\newtheorem{lemma}{Lemma}

\begin{document}
\title{Aleatoric uncertainty for Errors-in-Variables models in deep regression}

\author{J. Martin\thanks{Corresponding author. \Letter \, Jörg Martin: joerg.martin@ptb.de}  \thanks{Physikalisch-Technische Bundesanstalt, Abbestr. 2, 10587 Berlin} , C. Elster\footnotemark[2]}
\maketitle

\begin{abstract}
	A Bayesian treatment of deep learning allows for the computation of uncertainties associated with the predictions of deep neural networks. We show how the concept of Errors-in-Variables can be used in Bayesian deep regression to also account for the uncertainty associated with the input of the employed neural network. The presented approach thereby exploits a relevant, but generally overlooked, source of uncertainty and yields a decomposition of the predictive uncertainty into an aleatoric and epistemic part that is more complete and, in many cases, more consistent from a statistical perspective. We discuss the approach along various simulated and real examples and observe that using an Errors-in-Variables model leads to an increase in the uncertainty while preserving the prediction performance of models without Errors-in-Variables. For examples with known regression function we observe that this ground truth is substantially better covered by the Errors-in-Variables model, indicating that the presented approach leads to a more reliable uncertainty estimation. 
\end{abstract}

\section{Introduction}
\label{sec:introduction}

In recent years deep neural networks have proven to be a useful and powerful tool in various tasks, ranging from medical applications \cite{kooi2017large,kretz2019determination}, over language processing \cite{otter2020survey,kamath2019deep} to computer vision \cite{voulodimos2018deep,kendall2017uncertainties}, robotics \cite{pierson2017deep,lu2019rnn,li2021overview} and autonomous driving \cite{grigorescu2020survey,huang2020survey}.
In many applications, especially those in which reliability and safety are crucial \cite{mcallister2017concrete,litjens2017survey,sunderhauf2018limits}, it is valuable, if not indispensable, to know the uncertainty behind a prediction of a neural network. This work focuses on the uncertainty evaluation for neural networks that are trained for regression tasks \cite{gal2016dropout,kendall2017uncertainties,lakshminarayanan2016simple,maddox2019simple,schmahling2021framework,loquercio2020general}. Regression problems arise in a variety of areas \cite{kretz2019determination,hoffmann2021uncertainty,li2022diversified,martin2019application} and are typically given by a model $f_\theta$, parametrized by $\theta$, that links input data $x$ to outputs $y$\,:
\begin{align}
	\label{eq:classical_model}
	y = f_\theta(x) + \eps_y \,,
\end{align}
where $\eps_y\sim \Ncal(0, \sigma_y^2)$ is some normally distributed noise\footnote{The index $y$ should not be misunderstood as $\eps_y$ or $\sigma_y$ being $y$ dependent, but is supposed to indicate that the noise corresponds to the output. The same is true for $\eps_x$ and $\sigma_x$, which we introduce in \eqref{eq:new_model} below.} that disturbs the true label (i.e., value of the regression function) $f_\theta(x)$ corresponding to $x$. 

 \begin{figure}[t]
	\centering
	\includegraphics[width=0.8\textwidth]{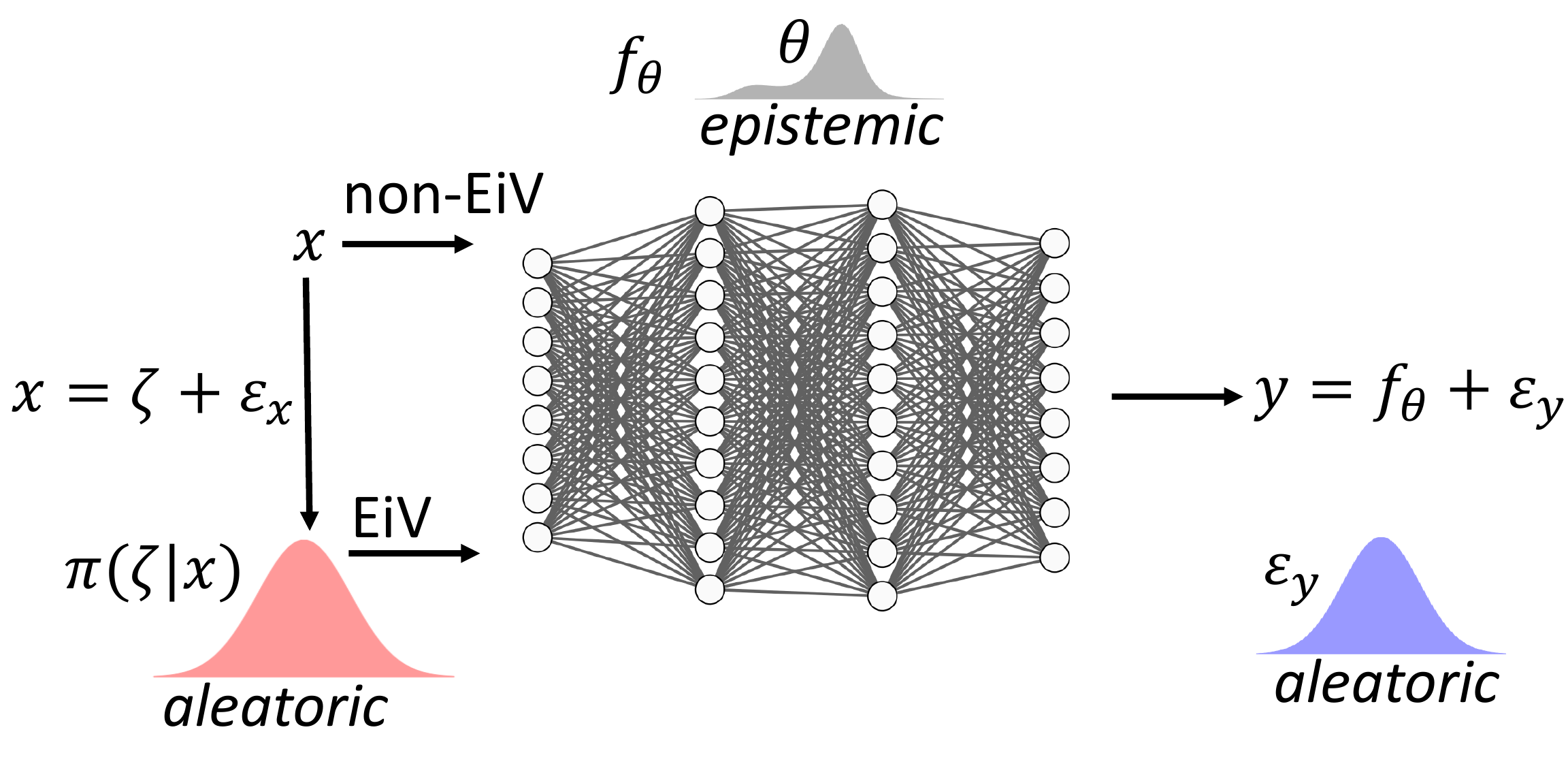}
	\caption{Illustration of the Errors-in-Variables (EiV) approach presented in this work and the method without EiV (non-EiV). The EiV model introduces an additional uncertainty to the input of the network that is aleatoric and, in contrast to the one linked with $y$, in many cases more coherent with the classical statistical view on uncertainty, cf. Section \ref{sec:introduction}. The network illustration was created using \cite{alexlenail}.} 
	\label{fig:illustration}
\end{figure}

In deep regression $f_\theta$ is a neural network with parameters $\theta$. Training this neural network means to infer a value for $\theta$ from pairs $(x,y)$ contained in a training set $\mathcal{D}$.  Uncertainties for predictions of a trained neural network are usually categorized using two different terms. \emph{Epistemic} uncertainty arises from the uncertainty about the trained model, that is about $\theta$ in the notation of \eqref{eq:classical_model}. In a Bayesian approach, as in this work, this uncertainty is described by the posterior $\pi(\theta|\mathcal{D})$, which is the distribution of $\theta$ conditional on the data $\mathcal{D}$ \cite{blundell2015weight,gal2016dropout,kingma2015variational}. This type of uncertainty vanishes as the number of observations tends to infinity, as follows for instance from the Bernstein-von Mises theorem. \emph{Aleatoric} uncertainty, on the other hand, describes an uncertainty that is inherent to the data and cannot be reduced even with an infinite training set. In the context of regression this corresponds to a noise such as $\eps_y$ in \eqref{eq:classical_model} and might be measured, for instance, using $\sigma_y$.
In deep learning, the aleatoric uncertainty expressed by $\sigma_y$ is often considered as important and is regarded as measuring inherent, irreducible label ambiguity \cite{depeweg2018decomposition, gustafsson2020evaluating, hullermeier2021aleatoric, kendall2017uncertainties}.
However, such a point of view no longer applies in those regression problems where the goal is to predict the regression function $f_\theta(x^*)$ at some (observed) $x^*$. In this case only the epistemic uncertainty about the network's parameter $\theta$ remains which, in principle, could become arbitrarily small as the training set grows.
However, in most cases $x^*=\zeta^*+\eps_x$ contains some noise $\eps_x$ as well, and one is actually interested in the value of $f_\theta(\zeta^\ast)$.
The uncertainty caused by the fact that $x^*$, and not $\zeta^*$, is observed then constitutes an aleatoric part of the uncertainty for the prediction of $f_\theta(\zeta^*)$ that is not covered by $\sigma_y$ nor by the uncertainty of $\theta$.

We here attempt to provide a more consistent view on these issues and, we argue, a more accurate depiction.
In many cases, it is a too crude assumption to presume that, while $y$ is deranged by noise, $x$ is not. 
Standard estimation procedures such as maximum likelihood or nonlinear least-squares become biased for a regression model of the form
\eqref{eq:classical_model} when $x$ is observed with noise \cite{fuller2009measurement}. Furthermore, a separate quantification of the aleatoric part of the uncertainty due to the noise of the input is not possible in a model like \eqref{eq:classical_model}, since the variance $\sigma_y^2$ accounts for noise in the output.
Following the idea of Errors-in-Variables (EiV), a quite classical concept in statistics \cite{fuller2009measurement}, we resolve these issues by changing \eqref{eq:classical_model} to
\begin{align}
	\label{eq:new_model}
	\begin{aligned}
		x &= \zeta + \eps_x \,,\qquad y = f_\theta(\zeta) + \eps_y \,,
	\end{aligned}
\end{align}
where $\zeta$ denotes the true, but unknown, input value. 
Besides allowing for noisy input data, model \eqref{eq:new_model} allows the aleatoric uncertainty to be treated in a manner that is more coherent from a statistical perspective. Throughout this work we will refer to \eqref{eq:new_model} as the \emph{EiV} model and to \eqref{eq:classical_model} as the \emph{non-EiV} model. An illustration of both approaches is given by Figure \ref{fig:illustration}.\nocite{alexlenail}

The uncertainty that arises in the EiV model for $\zeta$ given $x$ is an aleatoric uncertainty that must be taken into account whenever predicting $f_\theta(\zeta)$. It explains why there can be an uncertainty of the prediction even if there is no remaining epistemic uncertainty, without forcing this role on the output noise.

We will approach \eqref{eq:new_model} from a Bayesian point of view, which is a consistent way of describing uncertainty but one that involves the challenge of sampling from a high dimensional posterior on $\theta$. To overcome this obstacle, we build on the idea to use variational inference, as e.g. in \cite{blundell2015weight,gal2016dropout,gal2017concrete,kingma2015variational}. The main contributions of this work can be summarized as follows:

\begin{itemize}
	\item We show how variational inference for Bayesian neural networks can be combined with the concept of an uncertain input to construct a scalable Errors-in-Variables model for Bayesian deep learning.
	\item We show that for cases where the input is indeed uncertain, the treatment via an EiV model leads to a substantially improved coverage of the ground truth.
\end{itemize}

While the presented approach is in principle agnostic with regard to the specific form of the variational distribution, the results in Section \ref{sec:experiments} below were produced with the variational distribution induced by Monte Carlo dropout \cite{gal2016dropout}. Studying how well other methods perform under EiV could be an interesting subject of future work.

\subsubsection*{Existing work and structure of the article}

Errors-in-Variables is a statistical concept that has existed for decades \cite{gillard2006historical}. In \cite{van1998errors,van2000learning,bassu1999training,sragner2003improved,seghouane2001cost} the authors consider a non-Bayesian Errors-in-Variables framework for the training of neural networks, without the quantification of uncertainties. In \cite{wright1999bayesian,wright2000neural,pavone2018bayesian} the authors use a Laplace approximation to derive uncertainties for a neural network with uncertain inputs. Their approach requires however the computation of the inverse of the Hessian of the model w.r.t. network parameters which is computationally prohibitive for most modern neural networks. The same is true for methods based on Markov Chain Monte Carlo sampling \cite{zhang2011explicitly,yuan2020neural}. In \cite{xie2020input} input uncertainty is treated via a Gaussian process approximation to a neural network, which however requires, in theory, an infinite width of the network.

Many of the methods in the deep learning literature for uncertainty quantification that scale well with the number of parameters rely on variational inference \cite{blundell2015weight,gal2016dropout,gal2017concrete,kingma2015variational,duvenaud2016early,zhang2018noisy}. In this work, we build on the idea to use variational inference to obtain a scalable method for uncertainty quantification in deep learning, but elaborate this idea considerably by allowing for an uncertain input.

The article is structured as follows: in Section \ref{sec:an_eiv_model_for_deep_learning}, we discuss the generic approach presented in this work, give a proposal for the underlying priors and discuss some details useful for implementation. In Section \ref{sec:experiments}, we will discuss the results of some numerical experiments. The focus will be on simulated models with a known ground truth that allow the inferential behavior of the EiV model to be quantitatively assessed. Finally, we provide a discussion and some conclusions.

\section{An EiV model for deep learning}
\label{sec:an_eiv_model_for_deep_learning}

Suppose we have $N$ data points $\DD=\{(x_1,y_1), \ldots,(x_N,y_N)\}\subseteq \RR^{n_x}\times \RR^{n_y}$ that we model via 
\begin{align}
	\label{eq:EiVModel}
	\begin{aligned}
		x_i = \zeta_i + \eps_{x,i}\,,\qquad  y_i = f_\theta(\zeta_i) + \eps_{y,i} 
	\end{aligned}
\end{align}
with normally distributed $\eps_{x,i} \sim \Ncal(0,\sigma_x^2 I_{n_x \times n_x}), \,\eps_{y,i} \sim \Ncal(0,\sigma_y^2 I_{n_y \times n_y})$ and where the parameters $\theta \in \RR^p$ and $\zeta_1,\ldots,\zeta_N \in \RR^{n_x}$ are unknown. In this work, we will, similar to \cite{van1998errors,van2000learning}, fix $\sigma_x$ prior to training. For the output variance $\sigma_y$ we will use an initial estimate, that is updated during training. The function $f_\theta$ is a neural network. 
The $\zeta_i$ ought to be considered as the true, but unknown, inputs we would like to feed to $f_\theta$. Given $\zzeta=(\zeta_1,\ldots,\zeta_N),\,\theta$ and $\ssigma^2 =(\sigma_x^2, \sigma_y^2)$, the likelihood for the data $\mathcal{D}$ under \eqref{eq:EiVModel} is
\begin{align}
	\label{eq:eiv_likelihood}
	p(\DD|\theta,\zzeta, \ssigma^2) = \prod_{i=1}^N p(x_i|\zeta_i, \sigma_x^2) \cdot p(y_i|\theta,\zeta_i, \sigma_y^2) 
\end{align}
with $ p(x_i|\zeta_i, \sigma_x^2)=\Ncal(x_i|\zeta_i,\sigma_x^2 I_{n_x \times n_x}))$ and $p(y_i|\theta,\zeta_i, \sigma_y^2 I_{n_y \times n_y})=$\\$\Ncal(y_i|f_\theta(\zeta_i), \sigma_y^2 I_{n_y \times n_y})$.
While $\theta$ will be considered as the parameters of interest, the components of $\zzeta$ will be considered as nuisance parameters. Fixing a prior
\begin{align}
	\label{eq:theta_zzeta_prior}
	\pi(\theta,\zzeta) =\pi(\theta)\pi(\zzeta) = \pi(\theta) \prod_{i=1}^N \pi(\zeta_i)\,,
\end{align}
(cf. Section \ref{subsec:choosing_pitheta_pizeta_i_and_q_phi} below) we have, via Bayes' theorem,
\begin{align}
	\label{eq:Bayes}
	\pi(\theta|\DD, \ssigma^2) = \frac{\pi(\theta) \pi(\DD | \theta,\ssigma^2) }{\pi(\DD| \ssigma^2)}\,,
\end{align}
where $\pi(\DD| \ssigma^2) = \int \dd \theta \,\pi(\theta) \pi(\DD | \theta,\ssigma^2) $
and, due to \eqref{eq:eiv_likelihood} and Bayes' theorem,
\begin{align}
	\label{eq:eiv_marginal_likelihood}
	\begin{aligned}
		\pi(\DD | \theta,\ssigma^2) &= \int \dd \zzeta\,\pi(\zzeta)\, p(\DD|\theta,\zzeta,\ssigma^2)  \\ &= \prod_{i=1}^N  \pi(x_i|\sigma_x^2) \int \dd \zeta_i\, \pi(\zeta_i | x_i, \sigma_x^2) p(y_i|\zeta_i,\theta, \sigma_y^2) 
	\end{aligned}
\end{align}
with $\pi(x_i| \sigma_x^2)=\int \dd \zeta_i \pi(\zeta_i) p(x_i|\zeta_i, \sigma_x^2)$.  As we do not expect $\pi(\theta| \DD, \ssigma^2)$ to be feasible, we approximate it via a variational distribution
\begin{align}
	\label{eq:idea_var_inf}
	\pi(\theta|\DD, \ssigma^2) \approx q_\phi(\theta)
\end{align}
with a variational parameter $\phi$. In variational inference, the distance in \eqref{eq:idea_var_inf} is measured via the Kullback-Leibler divergence $\KL(q_\phi(\theta) \| \pi(\theta |\DD, \ssigma^2))$.
As we argue in Appendix \ref{sec:theoretical_aspects}, to find a $\phi$ that minimizes this divergence, we can use backpropagation on the following loss function
\begin{align}
	\label{eq:mcloss}
	\begin{aligned}
		\Lcal^{\mathrm{M.C.}}(\phi) &:=  -\frac{1}{M} \sum_{m=1}^M \log \left(\frac{1}{L} \sum_{l=1}^L p(y_{i_m}|\theta_m, \zeta_{i_m,l}, \sigma_y^2 ) \right) + \frac{1}{N} \KL(q_{\phi}(\theta)\|\pi(\theta)) 
	\end{aligned} 
\end{align}
where the minibatches $\{(x_{i_1}, y_{i_1}), \ldots, (x_{i_M}, y_{i_M})) \}$, the $M$ samples $\theta_m \sim q_\phi(\theta_m)$ and $M\cdot L$ samples $\zeta_{i_m, l} \sim \pi(\zeta_{i_m, l}|x_i, \sigma_x^2 I_{n_x \times n_x})$ are re-drawn in each optimization step.
We used throughout this work $M=1$ for training and $M=100$ for evaluation and $L=5$ for both, training and evaluation. 

Note that the way $\theta_m$ is sampled in \eqref{eq:mcloss} differs slightly from the way this is done in approaches such as \cite{gal2016dropout,gal2017concrete,kingma2015variational}, as $\theta_m$ is identical for all inputs $\zeta_{i_m,l}$ with the same $m$.\footnote{For the approach from \cite{kingma2015variational} this also means that Gaussian dropout of ``type A'' has to be used.}  The loss function in \eqref{eq:loss} and \eqref{eq:mcloss} is the backbone of the EiV algorithm. 
To make the first term in \eqref{eq:mcloss} numerically stable and suitable for backpropagation, the common "logsumexp" function can be used.
Before we can use \eqref{eq:mcloss} for training, however, we must fix the prior distributions and the variational distribution $q_\phi$.

\subsection{Choosing $\pi(\theta), \pi(\zeta)$ and $q_\phi$}
\label{subsec:choosing_pitheta_pizeta_i_and_q_phi}

Choosing $\pi(\theta)$ is rather standard in the literature on Bayesian neural networks and complies with choosing a regularization for $\theta$, cf. \cite{blundell2015weight}. We will here use the common choice of a centered normal distribution $\pi(\theta) = \Ncal(\theta | 0, \lambda_\theta^2 I_{p \times p})$. For Bernoulli dropout \cite{gal2016dropout} with rate $p$ the term $\frac{1}{N}\KL(q_{\phi}(\theta)\|\pi(\theta))$ in \eqref{eq:mcloss} then equals, up to a constant, $\frac{(1-p)}{2N\lambda_\theta^2}|\theta|^2$.

The prior $\pi(\zeta)$, used for each individual $\zeta_i$ in \eqref{eq:theta_zzeta_prior}, is more specific to the EiV approach and influences \eqref{eq:mcloss} through the posterior $\pi(\zeta|x,\sigma_x^2 I_{n_x \times n_x})$ from which we draw Monte Carlo samples for the first term. In this work we use an improper prior for $\zeta$ which leads to the posterior $\pi(\zeta|x,\sigma_x^2) = \mathcal{N}(\zeta|x,\sigma_x^2 I_{n_x \times n_x})$, cf. Lemma 1 in the Appendix \ref{subsec:variational_inference_for_lambda_zetarightarrow_infty}, that formalizes the variational inference applied in this work. For a proper, normally distributed $\pi(\zeta)$ the according posterior $\pi(\zeta|x,\sigma_x^2 I_{n_x \times n_x})$ is given in the Appendix \ref{subsec:informative_distribution_for_pizeta}.

The choice of $q_\phi$ is only limited by three requirements. First, we need to be able to sample from $q_\phi$. Second, we need an expression for the regularizer $\KL(q_{\phi}(\theta)\|\pi(\theta))$ - either explicitly or via Monte Carlo sampling \cite{kingma2013auto}. Finally, we have to be able to optimize the arising loss function $\Lcal^{\mathrm{M.C.}}(\phi)$ w.r.t. $\phi$ for example via the reparametrization trick \cite{kingma2013auto}. Merely for convenience we will restrict ourselves in this work to the popular choice of Monte Carlo dropout \cite{gal2016dropout} where $q_\phi$ arises from randomly dropping nodes of the network with some probability and where $\phi$ simply coincides with the network parameters. However, let us emphasize that the algorithm described in this work is by no means restricted to this particular choice but is usable for any $q_\phi$ for which the conditions above apply.

The full algorithm used for training the neural networks in this work is summarized in Algorithm \ref{alg:train}. Note, that we regularly update $\sigma_y$ during training to match the RMSE on the training data. 

\begin{algorithm}[t]
	\KwData{Training data $\DD=\{ (x_1,y_1), \ldots, (x_N,y_N) \}$}
	\KwResult{$\phi$ and $\sigma_y^2$}
	Fix $\pi(\zeta)$, $\pi(\theta)$, $\sigma_x$\;
	Fix $n_{\mathrm{train}},\,n_{\mathrm{update}\,\sigma_y}$\;
	Choose initial $\phi$ and $\sigma_y$\;
	\For{$j$ in $1,...,n_{\mbox{\scriptsize $\mathrm{train}$}}$}{
		\For{minibatches
$(x_{i_1},y_{i_1}), \ldots, (x_{i_M}, y_{i_M})$ from $\DD$\;
		}{
		Draw $\zeta_{i_m,1}, \ldots, \zeta_{i_m,L}$ from $\pi(\zeta_{i_{m,l}}|x_{i_m}, \sigma_x^2)$ for each $m$\;
		Draw $\theta_1,\ldots, \theta_M$ from $q_\phi(\theta_m)$\;
		Feedforward $(\zeta_{i_m,l})_{m\leq M,l\leq L}$ through $f_\theta$\;
		Compute $\Lcal^{\mathrm{M.C.}}(\phi)$ as in \eqref{eq:mcloss}\;
		Update $\phi$ via $\nabla_{\phi}\Lcal^{\mathrm{M.C.}}(\phi)$\;
		\uIf{$j$ multiple of $n_{\mathrm{update}\,\sigma_y}$}{
			Set $\sigma_y$ to RMSE on training data\;
		}
}
}
\caption{Training a neural network with EiV}
	\label{alg:train}
\end{algorithm}

\subsection{A new view on aleatoric uncertainty}
\label{subsec:a_new_view_on_aleatoric_uncertainty}
\label{subsec:computation_of_predictions_and_uncertainties}

Once the neural network has been trained the learned $\phi$ can be used to obtain uncertainties and predictions for a new $x^\ast$. Namely, we draw $\zeta^\ast$ from\footnote{Note that we understand $\zeta^\ast$ as independent of the $\zeta_i$ that generated the $x_i$ and $y_i$ in $\DD$, which is why we can drop the conditioning on $\DD$.} $\pi(\zeta^\ast|x^\ast, \sigma_x^2)=\pi(\zeta^\ast|x^\ast, \mathcal{D}, \sigma_x^2)$ and $\theta$ from $q_\phi(\theta)$ and use the distribution of the regression curve $f_\theta(\zeta^\ast)$. To express predictions and uncertainties, we can then either use quantiles or moments. For the latter, we set
\begin{align}
	\label{eq:EiV_pred_unc}
	m(x^\ast) =  \EE_{ \substack{\theta \sim \pi(\theta|\DD, \ssigma^2),\, \\\zeta^\ast \sim \pi(\zeta^\ast|x^\ast, \sigma_x^2)}}[f_\theta(\zeta^\ast)], \,
	u(x^\ast)^2 = \Var_{ \substack{\theta \sim \pi(\theta|\DD, \ssigma^2),\, \\\zeta^\ast \sim \pi(\zeta^\ast|x^\ast, \sigma_x^2)}}[f_\theta(\zeta^\ast)] \,.
\end{align}
By distinguishing between $\zeta^\ast$ and $x^\ast$ the EiV model introduces a new concept of prediction, namely $m$ in \eqref{eq:EiV_pred_unc}, that differs from the non-EiV model. The uncertainty $u(x^\ast)$ in \eqref{eq:EiV_pred_unc} can be split into (law of total variance)
\begin{align}
	\label{eq:EiV_uncertainty}
	\begin{aligned}
		u(x^\ast)^2 &= \EE_{\zeta^\ast \sim \pi(\zeta^\ast|x^\ast, \sigma_x^2)}(\Var_{ \theta \sim \pi(\theta|\DD, \ssigma^2)}[f_\theta(\zeta^\ast)]) \qquad \mbox{(epistemic)} \\
		       &+  \Var_{\zeta^\ast \sim \pi(\zeta^\ast|x^\ast, \sigma_x^2)}(\EE_{ \theta \sim \pi(\theta|\DD, \ssigma^2)}[f_\theta(\zeta^\ast)]) \,. \qquad \mbox{(aleatoric)}
	\end{aligned}	
\end{align}
The detailed algorithm on how to compute $m(x^\ast)$ and $u(x^\ast)$ is shown in Algorithm \ref{alg:eval}. 
\begin{algorithm}[t]
	\KwData{$x^\ast$}
	\KwResult{Approximation to $m(x^\ast)$ and $u(x^\ast)$}
	Sample $\zeta_1^\ast, \ldots, \zeta_L^\ast $ from $\pi(\zeta^\ast_l|x^\ast, \sigma_x^2)$\;
	Sample $\theta_1,\ldots, \theta_K$ from $q_\phi(\theta_k)$\;
	\For{$k$ in $1,\ldots,K$}{
		Feed-forward $m_{lk}=f_{\theta_k}(\zeta_l^\ast)$ for $l$ in $1,\ldots,L$\;
	}
	Set $m(x^\ast)=\frac{1}{L K} \sum_{l,k} m_{lk} $ \;
	Set $u(x^\ast)=\frac{1}{L K-1} \sqrt{\sum_{l,k} (m_{lk} - m(x^\ast))^2 }$ \;
	\caption{Prediction and uncertainty for a trained EiV model}
	\label{alg:eval}
\end{algorithm}
The ``classical'' treatment of epistemic and aleatoric uncertainty, which is based on the posterior predictive distribution and discussed in Section \ref{sec:introduction}, can be easily combined with the above. For the posterior predictive distribution $\pi(y^\ast|x^\ast, \mathcal{D}, \ssigma^2)$, we draw, for each sample $f_\theta(\zeta^\ast)$, labels $y^\ast$ from $\mathcal{N}(f_\theta(\zeta^\ast), \sigma_y^2 I_{n_y \times n_y})$. The variance of this distribution, also known as the \emph{total uncertainty}, can then be split into
\begin{align}
	\label{eq:total_unc}
	\Var_{y^\ast \sim \pi(y^\ast|x^\ast,\DD, \ssigma^2)} [y^\ast] = u(x^\ast)^2 +\sigma_y^2
\end{align}
and is therefore simply augmented by an extra term $\sigma_y^2$. For $\sigma_x \searrow 0$ the aleatoric part of $u(x^\ast)$ in \eqref{eq:EiV_uncertainty} vanishes, $u(x^\ast)$ coincides with the epistemic uncertainty and \eqref{eq:total_unc} morphs into the conventional split of aleatoric and epistemic uncertainty.  

The usage of the aleatoric part $\sigma_y^2$ that appears in \eqref{eq:total_unc} as an uncertainty is only justified if one is really interested in the uncertainty of a noise-perturbed label $y^\ast$ given $x^\ast$. If one is actually interested in the value of the regression function $f_\theta(\zeta^\ast)$ (EiV) or $f_\theta(x^\ast)$ (non-EiV), it is inappropriate to take the second term in \eqref{eq:total_unc} into account. By introducing an additional aleatoric uncertainty in \eqref{eq:EiV_uncertainty}, the EiV model introduces an aleatoric uncertainty that is still present in such cases. In summary, the usage of the second term of \eqref{eq:EiV_uncertainty} as an additional aleatoric uncertainty has several advantages
\begin{itemize}
	\item This aleatoric uncertainty is still available if one is interested in the prediction $f_\theta(\zeta)$ and will, in particular, not vanish for large training data sets.
	\item It exploits a source of uncertainty that is not present in the non-EiV model and thus gives a more complete description. 
	\item As we will observe in Section \ref{sec:experiments} below, this enhanced uncertainty is necessary to achieve a sufficient coverage of the ground truth when the input to the neural network is uncertain. 
\end{itemize}

\section{Experiments}
\label{sec:experiments}

In this section, we compare the performance of EiV models with corresponding non-EiV models on various data sets. To this end we use 

\begin{itemize}
	\item 4 simulated data sets with known ground truth $g:\zeta \mapsto g(\zeta)$. For illustration purposes all of the chosen problems are one-dimensional in in- and output. We study three polynomial problems: a linear, a quadratic and a cubic model \cite{hernandez2015probabilistic,lakshminarayanan2016simple}. We will refer to these data sets as \emph{linear}, \emph{quadratic} and \emph{cubic} in the following. In addition we study a sinusoidal problem, as in \cite{blundell2015weight}, to which we will refer as \emph{sine}. In contrast to \cite{hernandez2015probabilistic,lakshminarayanan2016simple,blundell2015weight} we used noisy modifications of the input variables for training as in \eqref{eq:new_model}. Details on the problems can be found in Appendix \ref{sec:details_on_training_data}.
	\item 9 real data sets, with an unknown ground truth, that were used before, e.g. in \cite{hernandez2015probabilistic, gal2016dropout, lakshminarayanan2016simple}. All real data sets were normalized to have mean 0 and standard deviation 1 in each feature and label dimension.
\end{itemize}

For all data sets we trained fully connected neural networks with 4 hidden layers and dropout layers after each hidden layer. The detailed architecture is sketched in the Appendix \ref{sec:network_and_training_details}. We used two different loss functions: the standard MC dropout loss function from \cite{gal2016dropout} and the EiV modification we proposed in Section \ref{sec:an_eiv_model_for_deep_learning}. For each loss function and each data set, we trained 10 different neural networks using 10 different random seeds\footnote{For each run, we used a different split of the full data into training and test set.}. For both approaches, EiV and non-EiV, we used the same training hyperparameters such as epoch number and learning rate. These parameters differ between data sets and are listed in Table \ref{tab:network_and_training} in the appendix. In particular, we used different values of $\sigma_x$, once more listed in Table \ref{tab:network_and_training}. For the simulated data set we used the same value than the one that was used for generating the data. For all real data sets we used $\sigma_x=0.05$, except for the ``naval propulsion'' data set where we found that that this choice leads to an intolerably high root mean squared error (RMSE), so that we chose $\sigma_x = 0.025$ instead. For further details on the training we refer to Section \ref{sec:network_and_training_details} in the appendix. 

\begin{figure}[t]\centering
	\begin{subfigure}[t]{0.50\textwidth}
		\centering
		\includegraphics[width=\textwidth]{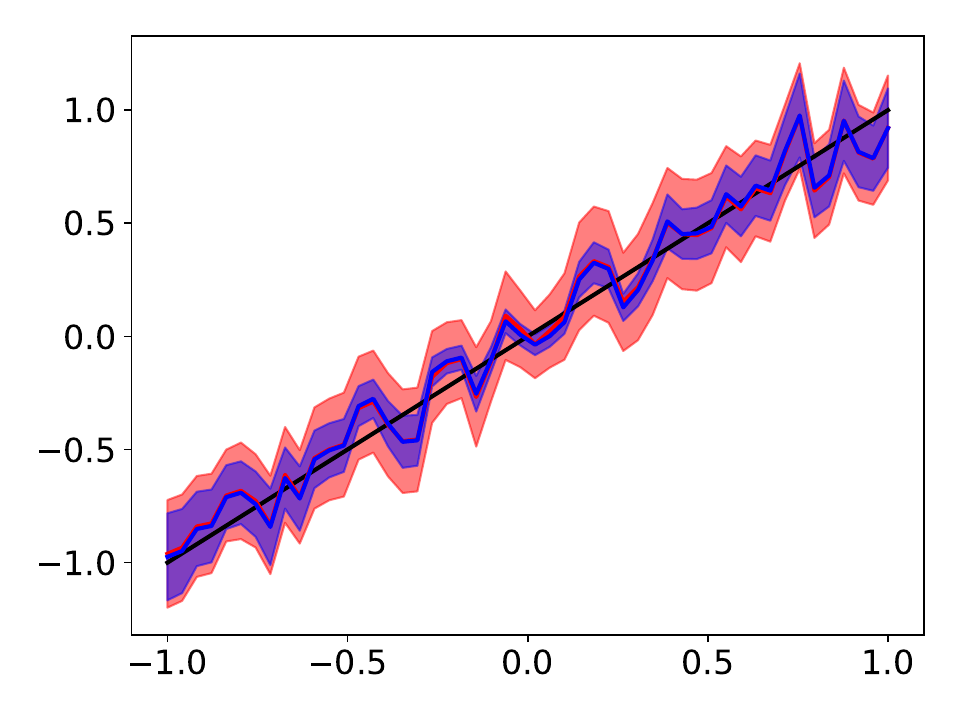}
		\caption{Prediction EiV and non-EiV}
		\label{subfig:prediction_linear}
	\end{subfigure}%
	~
	\begin{subfigure}[t]{0.50\textwidth}
		\centering
		\includegraphics[width=\textwidth]{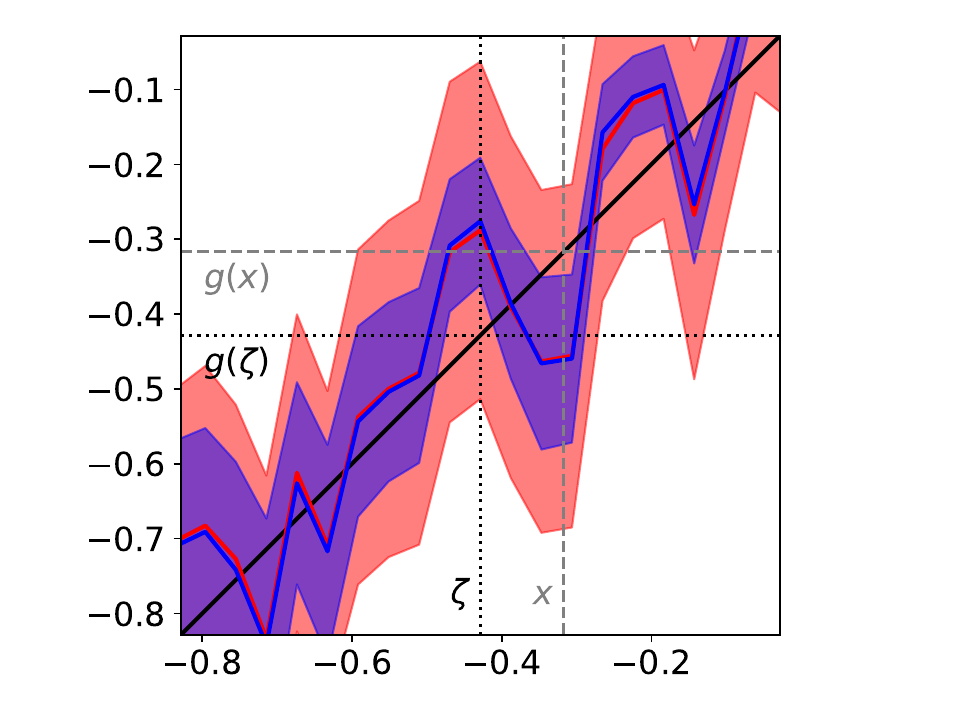}
		\caption{cut-out of Figure \ref{subfig:prediction_linear}}
		\label{subfig:prediction_linear_zoom}
	\end{subfigure}
	\caption{Prediction of the EiV model (red solid line) and non-EiV model (blue solid line) for the \emph{linear} data set (cf. Section \ref{sec:experiments}) together with their uncertainties (shaded areas) times 1.96. The ground truth $g: \zeta \mapsto \zeta$ underlying the data set is depicted by the black solid line. The right hand side shows a cut-out of the left hand side, together with an example of a $\zeta$ (black dotted vertical line) and the corresponding $x \sim p(x|\zeta,\sigma_x^2)$ (gray dashed vertical line) used for the evaluation of the two models. The corresponding values $g(\zeta)$ and $g(x)$ are marked by the two horizontal lines.}
	\label{fig:prediction_linear}
\end{figure}

Figure \ref{subfig:prediction_linear} shows the prediction of the EiV model (red) and non-EiV model (blue) trained on the linear data set, mentioned above, that uses the ground truth $g(\zeta):\, \zeta \mapsto \zeta$ (black). For 50 equidistant $\zeta$ in the range $[-1,1]$ of the training data we drew $x\sim p(x|\zeta, \sigma_x^2)$ and then computed, as predictions, $\EE_{\theta \sim q_\phi(\theta)}[f_\theta(x)]$ for the non-EiV model and $m(x)$ as in \eqref{eq:EiV_pred_unc} for the EiV model. To even out random effects due to the training we averaged all predictions over the 10 different training runs and plotted the results as blue (EiV) and red (non-EiV) lines in Figure \ref{subfig:prediction_linear} against $\zeta$. The corresponding values of $g(\zeta)$ are shown by the black line. The shaded areas mark the epistemic uncertainty $\left(\Var_{\theta \sim q_\phi(\theta)}(f_\theta(x))\right)^{1/2}$ for the non-EiV model (blue) and the combined uncertainty $u(x)$ from \eqref{eq:EiV_pred_unc} for the EiV model (red), both again averaged over the 10 training runs. 

\begin{figure}[t]
	\centering
	\includegraphics[width=0.6\textwidth]{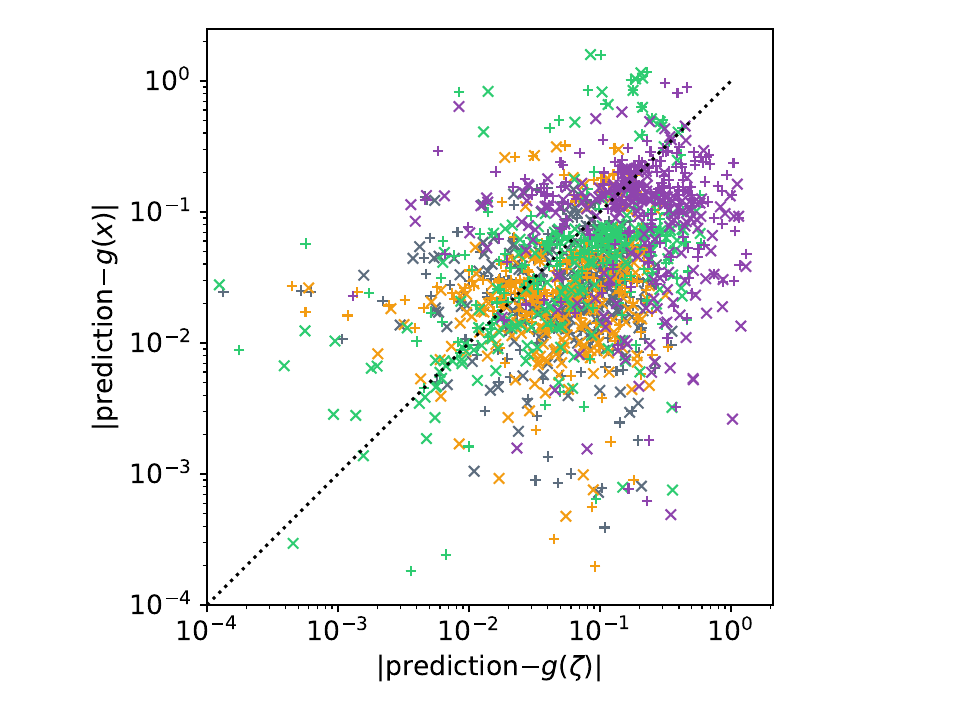}
	\caption{The deviation of the prediction of the EiV model (marker $+$) and the non-EiV model (marker $\times$) from $g(\zeta)$ (abscissa) and $g(x)$ (ordinate), where $x\sim p(x|\zeta,\sigma_x^2)$, for all simulated data sets in this work: \emph{linear} (gray), \emph{quadratic} (orange), \emph{cubic} (green) and \emph{sine} (purple). The dotted black line marks the diagonal.}
	\label{fig:deviation_scatter}
\end{figure}
Figure \ref{subfig:prediction_linear} shows a pattern which we observe for all data sets studied in this work: the predictions of both models are similar, but the uncertainties differ markedly. Moreover, the ground truth (black line) is substantially better covered by the uncertainties of the EiV model. To get better insight in these observations, Figure \ref{subfig:prediction_linear_zoom} shows a cut-out of Figure \ref{subfig:prediction_linear} together with a single choice of $\zeta$ (vertical dotted black line) and the corresponding draw $x\sim p(x|\zeta, \sigma_x^2)$ (vertical dashed gray line) used for the prediction. The two horizontal lines show the values of $g(\zeta)$ (black dotted) and $g(x)$ (gray dashed). Apparently the prediction of both, the EiV and non-EiV model, is considerably closer to $g(x)$ than to $g(\zeta)$ .
The epistemic, i.e. parameter, uncertainty of the non-EiV model is not large enough to account for the deviation $g(\zeta)-g(x)$ and does therefore not cover the ground truth $g(\zeta)$. Through $\pi(\zeta | x) = \Ncal(\zeta|x,\sigma_x^2 I_{n_x \times n_x})$ (cf. Section \ref{subsec:choosing_pitheta_pizeta_i_and_q_phi}) the EiV model accounts for the uncertainty about $\zeta$ and thereby yields an uncertainty $u(x)$ that covers the ground truth $g(\zeta)$.

\begin{figure}[t]\centering
	\begin{subfigure}[t]{0.32\textwidth}
		\centering
		\includegraphics[width=\textwidth]{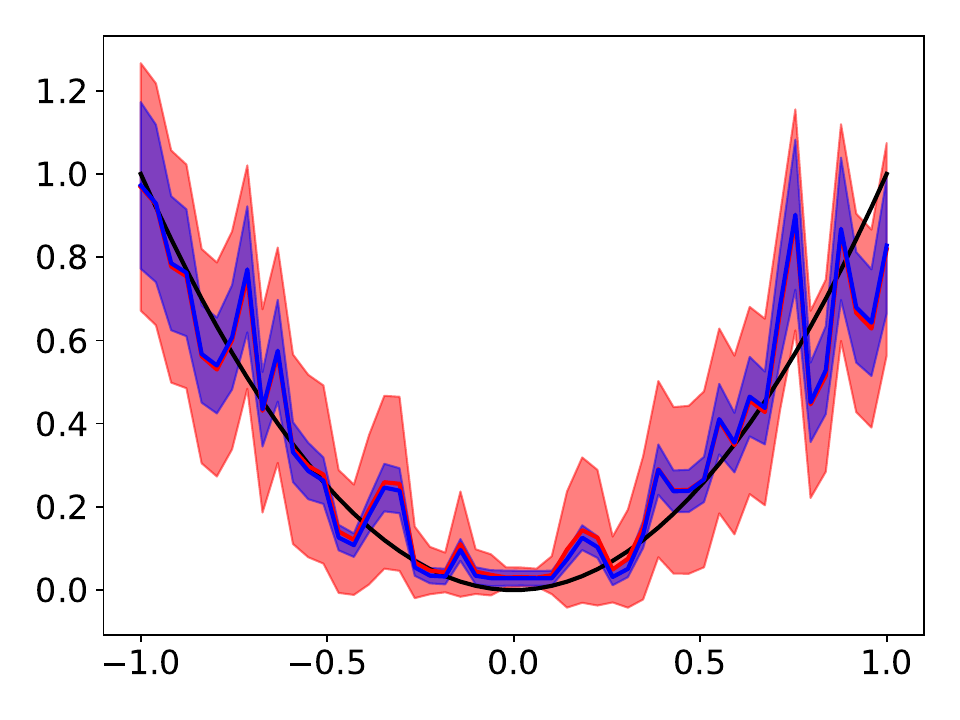}
		\caption{quadratic}
		\label{subfig:quadratic}
	\end{subfigure}%
	~
	\begin{subfigure}[t]{0.32\textwidth}
		\centering
		\includegraphics[width=\textwidth]{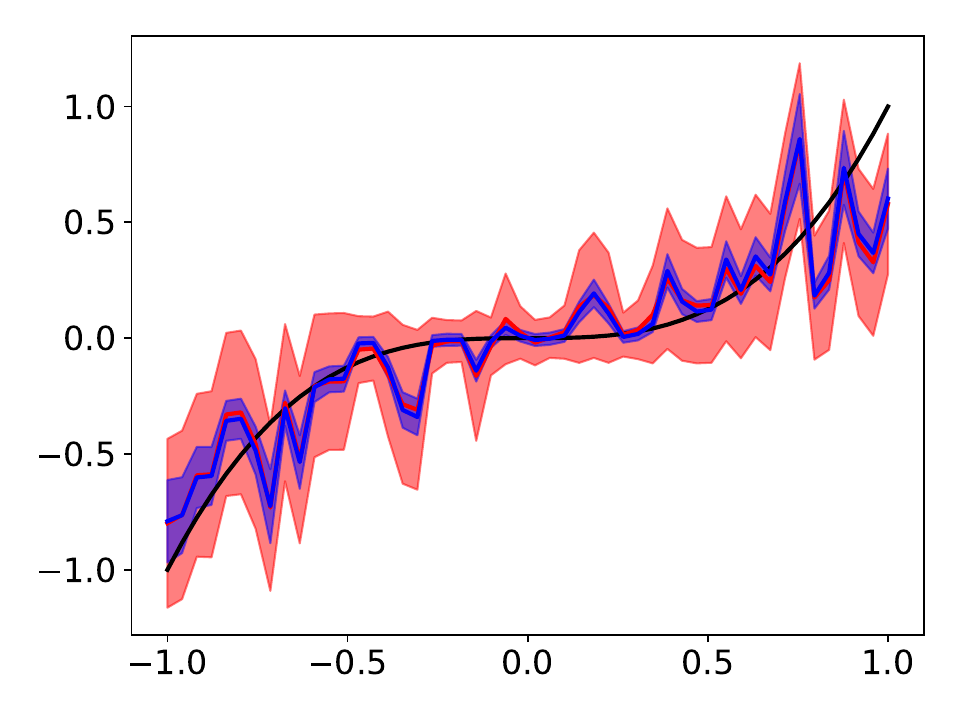}
		\caption{cubic}
		\label{subfig:cubic}
	\end{subfigure}%
	~
	\begin{subfigure}[t]{0.32\textwidth}
		\centering
		\includegraphics[width=\textwidth]{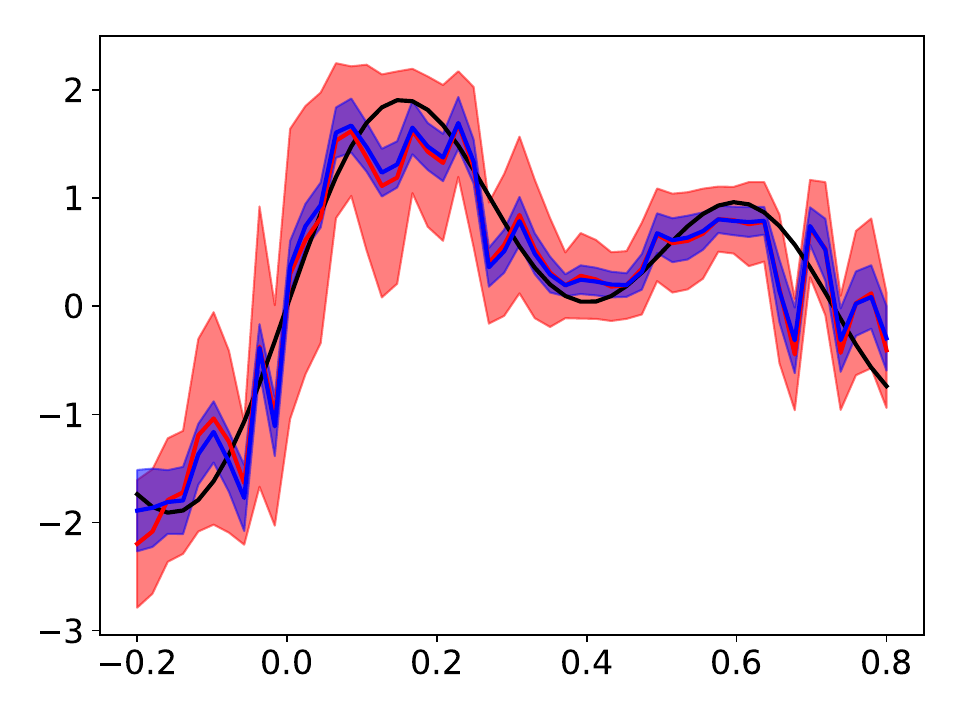}
		\caption{sine}
		\label{subfig:sine}
	\end{subfigure}
	\caption{Prediction of the EiV model (red) and the non-EiV model (blue) for three simulated data sets, together with their uncertainties (times 1.96) depicted by the red (EiV) and blue (non-EiV) area. The used data sets are the \emph{quadratic} (left), \emph{cubic} (middle) and \emph{sine} (right) data set, described in Section \ref{sec:experiments}. The corresponding ground truth is given by the black solid line.}
	\label{fig:other_predictions}
\end{figure}

\begin{figure}[t]\centering
	\begin{subfigure}[t]{0.33\textwidth}
		\centering
		\includegraphics[width=\textwidth]{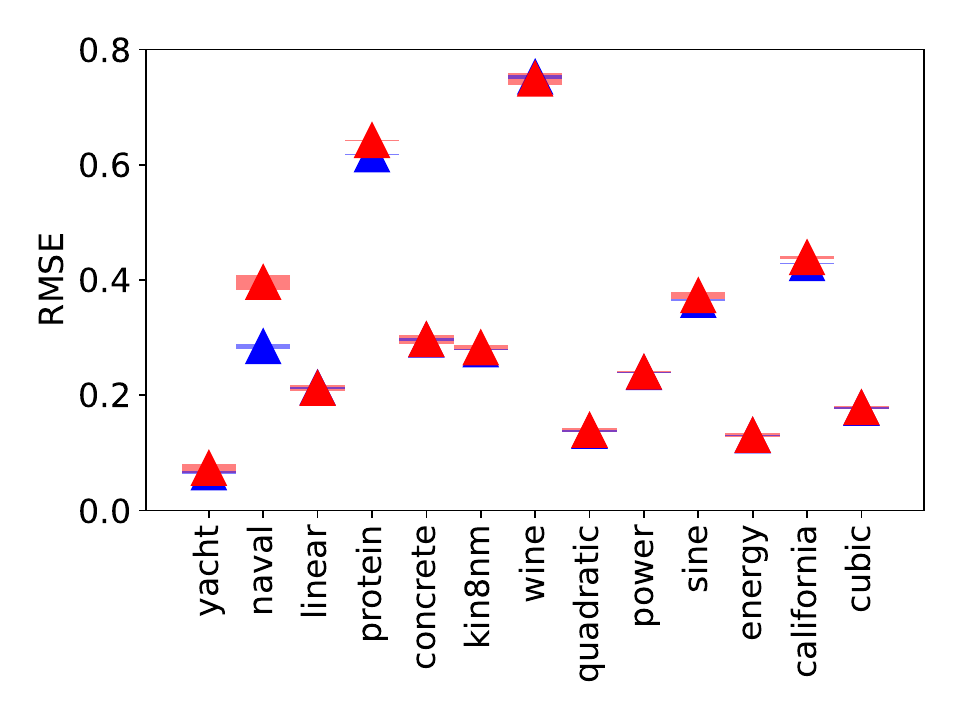}
		\caption{RMSE for all data sets}
		\label{subfig:RMSE_bar}
	\end{subfigure}%
	~
	\begin{subfigure}[t]{0.33\textwidth}
		\centering
		\includegraphics[width=\textwidth]{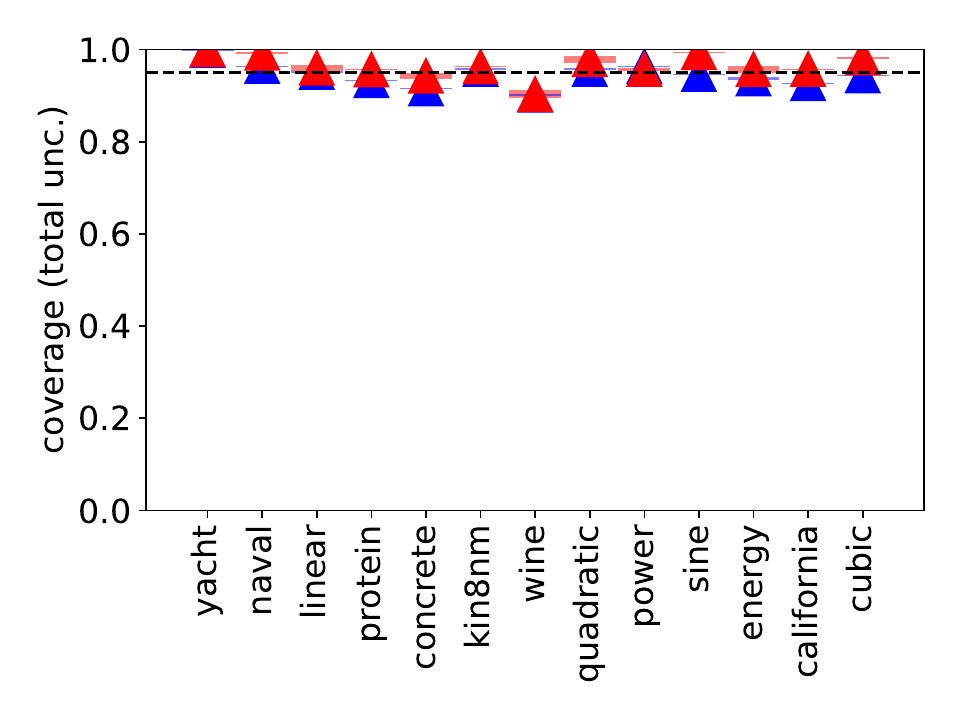}
		\caption{Coverage (total unc.)}
		\label{subfig:total_cov_bar}
	\end{subfigure}%
	~
	\begin{subfigure}[t]{0.33\textwidth}
		\centering
		\includegraphics[width=\textwidth]{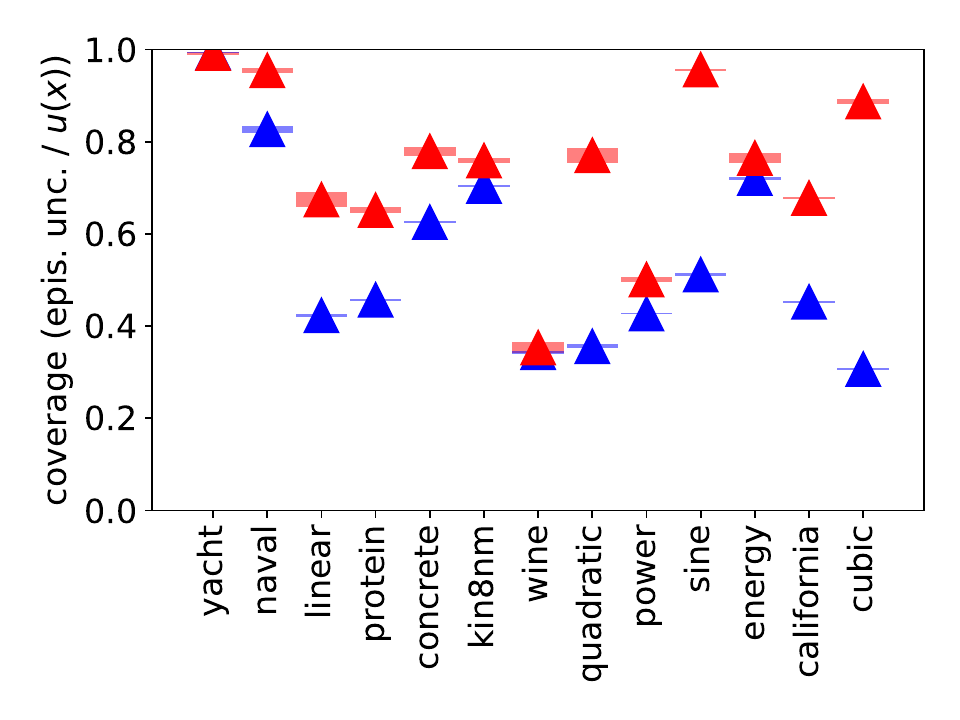}
		\caption{Coverage (epis. unc./$u$)}
		\label{subfig:noisy_cov_bar}
	\end{subfigure}
	\caption{The RMSE (left) and the coverage of the labels $y$ by the total uncertainty times 1.96 (middle) and the epistemic uncertainty (non-EiV) or $u$ (EiV) times 1.96 (right) for all data sets used in this work. Results from the non-EiV model are shown in blue, results from the EiV model are shown in red. The dashed line in the middle plot depicts 0.95. The shown results were averaged over 10 different training runs. The bars surrounding the markers depict the standard errors from these 10 runs.}
	\label{fig:rmse_and_noisy_coverage}
\end{figure}

Figure \ref{fig:deviation_scatter} shows that this sort of behavior is systematic. For each simulated data set, linear (gray), quadratic (orange), cubic (green) and sine (purple),
and their corresponding ground truth function $g$ we drew 200 test $\zeta$ and corresponding $x\sim p(x|\zeta, \sigma_x^2)$ and computed the deviation of the prediction of the EiV (marker $+$) and the non-EiV model (marker $\times$) from $g(\zeta)$ (x-axis) and $g(x)$ (y-axis)\footnote{As above, all predictions of the networks obtained from different training runs where averaged before computing the deviations.}. For all considered data sets, most of the points are located below the diagonal (dashed black line). In other words, for the majority of pairs $(x,\zeta)$ the prediction of both models, EiV and non-EiV, is closer to $g(x)$ than to $g(\zeta)$. As $g(x)$ usually differs from the ground truth $g(\zeta)$ this leads to an error.

\begin{figure}[t]\centering
	\begin{subfigure}[t]{0.49\textwidth}
		\centering
		\includegraphics[width=\textwidth]{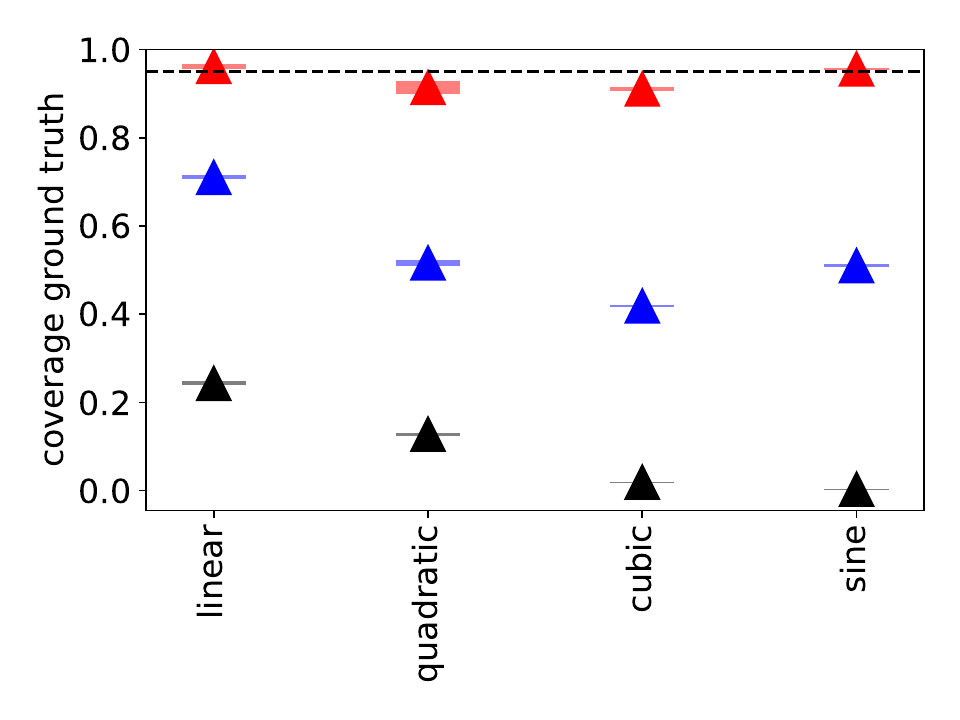}
		\caption{Coverage of the ground truth for simulated data}
		\label{subfig:true_cov_bar}
	\end{subfigure}%
	~
	\begin{subfigure}[t]{0.49\textwidth}
		\centering
		\includegraphics[width=\textwidth]{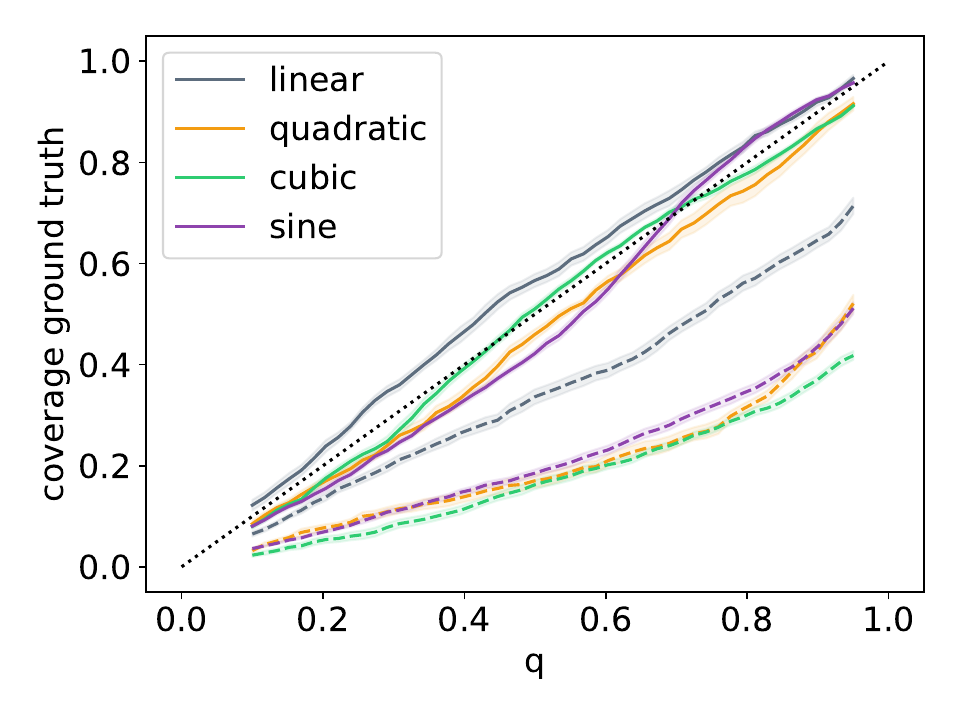}
		\caption{Coverage of the ground truth for various uncertainty intervals}
		\label{subfig:true_cov_vs_q}
	\end{subfigure}
	\caption{\textit{Left}: The coverage of the ground truth by $1.96\cdot u$ for a EiV model (red) or $1.96$ times the epistemic uncertainty for a non-EiV model (blue) and a linear Bayesian regression model (black) for the simulated data sets used in this work. \textit{Right}: The coverage of the ground truth for various intervals, determined by $q$ as described in Section \ref{sec:experiments}. EiV coverages are depicted by solid lines, whereas the non-EiV coverages are depicted by dashed lines. In theory, the coverages should coincide with $q$, shown by the black, dotted diagonal.}
	\label{fig:true_coverage}
\end{figure}

We already saw in Figure \ref{fig:prediction_linear} that for the linear data set this error is not sufficiently covered by the non-EiV model but is well covered by the EiV approach. Figure \ref{fig:other_predictions} shows that this is also true for the other simulated data sets considered in this work. We observe, once more, that the predictions of both models are quite similar whereas the uncertainty is substantially increased for the EiV model. The ground truth $g:\,\zeta\mapsto g(\zeta)$ for all three data sets (black solid line) is substantially better covered by the EiV model. We will give full range coverages for all four simulated data sets in Figure \ref{fig:true_coverage} below and observe that they match well the theoretical expectation.

Figure \ref{fig:rmse_and_noisy_coverage} shows the performance of the EiV and non-EiV model for all data sets considered in this work, including the 9 real data sets mentioned above. Figure \ref{subfig:RMSE_bar} shows the root-mean-squared-error\footnote{The RMSE was computed w.r.t the labels $y$, even for the simulated models with known ground truth $g$.} (RMSE) of the EiV models (red) and non-EiV models (blue). Again, all results were averaged over 10 training runs. The corresponding standard errors (of the mean) are marked by the small bars surrounding the markers. We observe, with the only exception of the naval data set, an almost identical RMSE for both models. This is in consistence with the observation from Figure \ref{fig:prediction_linear} and \ref{fig:other_predictions} where the predictions of both models, and thus the RMSE, are pretty similar. Figure \ref{subfig:total_cov_bar} shows the coverage of the labels $y$ within the test set by the \emph{total} uncertainty (times 1.96) of both models. 
The vertical dashed line shows 0.95, which is the value one would expect for a normal distribution. Both models, EiV and non-EiV, show similar coverages which are close to this optimal value. When looking at the coverage by $u$ as in \eqref{eq:EiV_pred_unc} (for EiV) and the epistemic uncertainty  (for non-EiV) we see, however, a substantial difference between both models. As in Figure \ref{fig:prediction_linear} and \ref{fig:other_predictions}, we observe a substantially increased coverage through introducing Errors-in-Variables. 
While this backs our proposed new view on the impact of aleatoric input uncertainty - explaining some of the error via an aleatoric input uncertainty and less by the aleatoric output uncertainty - Figure \ref{subfig:true_cov_bar} does not allow for a judgment which of the two models is the better description of ``reality'', as we have no access to an optimal coverage value as in Figure \ref{subfig:total_cov_bar}.

To allow for such a judgment we have to restrict ourselves, once more, to those cases where we have access to the ground truth $g:\,\zeta\mapsto g(\zeta)$, in other words to the simulated data sets. Figure \ref{subfig:true_cov_bar} shows the coverage of the ground truth, that is the values $g(\zeta)$, for the test points of the 4 simulated data sets, that are covered by (an interval around the estimate of width) 1.96 times the epistemic uncertainty (for non-EiV, in blue) or $u$ (for EiV, in red).  The optimal value of 0.95 is indicated by the dashed, black line. In addition, we plotted the coverage of a Bayesian regression model (black) with an noninformative prior and a regression function based on the $g$ used for generating the data \cite{schmahling2021framework}, cf. Appendix \ref{subsec:simulated_data} for details. 
Results were, once more, averaged over 10 training runs with standard errors plotted by the bars next to the markers.
We observe that for all 4 data sets, the EiV approach yields a coverage of the ground truth that is far closer to the optimal value of 0.95 than the coverage achieved by the non-EiV model and the Bayesian regression model based on $g$. The insufficient coverage of the latter reveals that the low coverage of the non-EiV model is not based on a model misfit but really on an underestimated uncertainty: for both, the non-EiV model and the $g$ based model, the estimated uncertainty is not sufficient to cover the error that arises from the fact that we can only use $x$ and not $\zeta$ as an input.

One might argue that the better coverage of the EiV model could not root in a better description of ``reality'' by the EiV model but solely in the fact that the additional aleatoric input uncertainty simply raises the coverage closer to the maximal coverage of 1.0 and that 0.95 is close to the latter. Figure \ref{subfig:true_cov_vs_q} shows that this is not the case. For various values of $q$ between 0 and 1 we computed the corresponding multiple of the epistemic uncertainty (for non-EiV) and $u$ (for EiV) that should, in theory, lead to a coverage of $q$. On the ordinate we plotted the actual coverage of the two methods for the 4 simulated data sets. The results for the EiV method (solid lines) and the non-EiV methods (dashed) were averaged over the 10 training runs. The shaded area depicts the corresponding standard error. For all values of $q$ we observe that the EiV model is substantially closer to the theoretical value of $q$, depicted by the diagonal (dotted, black). This indicates that the uncertainty of the EiV model describes the error of its prediction far more reliable.

Let us summarize our observations:

\begin{itemize}
	\item EiV and non-EiV yield comparable predictions and thus show a comparable RMSE for both, simulated and real, data sets. The coverage of noisy labels through the total uncertainty is also comparable for both models and close to the optimal value.
	\item Using EiV leads to an increased coverage of both, the ground truth and noisy labels, when not including the aleatoric output uncertainty.
	\item This indicates that EiV relies less on the aleatoric output uncertainty to explain its error. This is achieved by introducing an aleatoric \emph{input} uncertainty.
	\item In cases where we have access to the ground truth the increase in coverage of the ground truth by the EiV model matches substantially better the theoretical expectation and is thus a more reliable description of the problem.
\end{itemize}

\section{Discussion and outlook}
\label{sec:discussion_and_outlook}

This article studies the effect of using an EiV model in Bayesian deep learning. As a posterior distribution for the true (but usually unknown) input $\zeta$ we derive in Section \ref{sec:an_eiv_model_for_deep_learning} the posterior distribution $\pi(\zeta|x) = \Ncal(\zeta|x,\sigma_x^2  I_{n_x \times n_x})$ which, loosely speaking, takes the observed $x$ as an estimate for $\zeta$ but equips it with an uncertainty. In \cite{van1998errors,van2000learning} the authors used a non-Bayesian Errors-in-Variables approach and constructed a guess on $\zeta$ via updating it through backpropagation. They observe that this reduces the bias of the estimated network parameters, at least if repeated measurements of $x$ for each $\zeta$ are available (which is rarely the case for most deep learning applications). Translating such an approach to a Bayesian approach, as in this work, would complicate the presented framework, raise the computational burden and, as argued in \cite{van1998errors}, raise the need for additional precautions, such as early stopping, to prevent the EiV approach from overfitting. However, such a study could be an exciting outlook for future work on Bayesian EiV in deep learning.

For the experiments in Section \ref{sec:experiments} we fixed, similar as in \cite{van1998errors,van2000learning}, $\sigma_x$ prior to training. In those cases where one has access to several $x$ for a fixed $\zeta$ the value of $\sigma_x$ can be estimated from the data. In all other cases $\sigma_x$ has to be estimated from prior knowledge or by a reasonable guess. We observed that learning $\sigma_x$ during training via \eqref{eq:mcloss} leads to overfitting, that is the learned $\sigma_x$ is pushed towards 0 during training. A compromise that takes prior knowledge into account but still allows for an adaption of $\sigma_x$ during training could be a modification of the presented approach that is based on Bayesian hierarchical modeling. We will leave such considerations to future work.

To keep the setup simple we restricted ourselves to variational inference based on Bernoulli dropout as in \cite{gal2016dropout,kendall2017uncertainties} and to regression tasks. Analyzing how different approaches perform under Errors-in-Variables would be a natural follow-up study to this article. The same is true for an enhancement of the distributions involved in \eqref{eq:new_model}, which could, for instance, involve an anisotropic or heteroscedastic adaptation or the usage of a learnable push-forward mapping.

Finally, while we studied in this work both, real and simulated data, only the simulated cases allow us to evaluate the quality of the uncertainty $u$ in a conclusive manner. The difficulty of assessing uncertainties without access to a ground truth is a well-known problem \cite{hullermeier2021aleatoric,schmahling2021framework}, whose solution is well beyond the scope of this article. A future work that would deepen the understanding of the performance for data without a ground truth could involve a study, again on simulated data, about the sensitivity of the EiV uncertainty quantification to deviations between the model assumptions in \eqref{eq:EiVModel} and the generation of the data, e.g. with respect to $\sigma_x$ or normally distributed noise.

\section{Conclusion}
\label{sec:conclusion}

In this work, we have shown how Errors-in-Variables (EiV), a classical concept from statistics, can be combined with existing Bayesian methods for uncertainty quantification in deep regression. This not only allows to treat the input of the network as equipped with an uncertainty but also provides a notion of aleatoric uncertainty that is, in many cases, more coherent with statistics. 
We found this method to give similar predictions to those of the non-EiV method but with an increased uncertainty. For examples with known ground truth and noisy inputs, this increased uncertainty was observed to be necessary, in order to achieve a sufficient coverage of the regression function, which indicates that using an EiV model leads to a more robust and reliable uncertainty quantification in applications where uncertain inputs are considered.

\section*{Acknowledgment}

This work was created within the M4AIM project.

\section*{Compliance with Ethical Standards}

The authors of this work are not aware of any conflict of interest.

\bibliographystyle{abbrv}
\bibliography{eiv_dl}
\appendix

\section{Theoretical aspects}
\label{sec:theoretical_aspects}

\subsection{Variational inference for informative $\pi(\zeta)$}
\label{subsec:variational_inference_for_informative_pizeta}

Given the setup from Section \ref{sec:an_eiv_model_for_deep_learning} the Kullback-Leibler divergence between the variational distribution $q_\phi(\theta)$ and the posterior $\pi(\theta |\DD, \ssigma^2))$ from \eqref{eq:Bayes} can be written as 

\begin{align*}
\KL(q_\phi(\theta) \| \pi(\theta |\DD, \ssigma^2)) = -\sum_{i=1}^{N}\int\dd\theta\,q_{\phi}(\theta)\log\left(\int\dd\zeta_{i}\pi(\zeta_{i}|x_{i},\sigma_x^2)p(y_{i}|\theta,\zeta_{i}, \sigma_y^2)\right)\\+\KL(q_{\phi}(\theta)\|\pi(\theta))-\sum_{i=1}^{N}\log\pi(x_{i}|\sigma_x^2) + \log\pi(\DD|\ssigma^2) \,,
\end{align*}
where we used \eqref{eq:Bayes}. Dropping the last two terms, which are independent of $\phi$, we arrive at the following loss function
\begin{align}
	\label{eq:loss}
\begin{aligned}
	\Lcal(\phi) = -\sum_{i=1}^{N}\int\dd\theta\,q_{\phi}(\theta)\log\left(\int\dd\zeta_{i}\pi(\zeta_{i}|x_{i},\sigma_x^2)p(y_{i}|\theta,\zeta_{i}, \sigma_y^2)\right)\\+\KL(q_{\phi}(\theta)\|\pi(\theta)) \,.
\end{aligned} 
\end{align}
Finding the $\phi$ that minimizes $\pi(\theta |\DD, \ssigma^2))$ is therefore equivalent to the following minimization problem
\begin{align}
	\label{eq:argmin_loss}
	\phi = \argmin_{\phi} \Lcal(\phi)\,.
\end{align}

In practice, we can use the Monte Carlo approximation \eqref{eq:mcloss} instead.

\subsection{Informative distribution for $\pi(\zeta)$}
\label{subsec:informative_distribution_for_pizeta}

For the choice $\pi(\zeta) = \Ncal(\zeta | 0, \lambda_\zeta^2 I_{n_x \times n_x})$ we obtain, for the setup of Section \ref{sec:an_eiv_model_for_deep_learning}, the following distributions via standard Bayesian calculus \cite{robert2007bayesian}
\begin{align}
	\label{eq:posterior_zeta}
\begin{aligned}
	\pi(\zeta| x, \sigma_x^2) &= \Ncal\Big( \zeta \Big| (1+\frac{\sigma_x^2}{\lambda_\zeta^2})^{-1} x, (1+\frac{\sigma_x^2}{\lambda_\zeta^2})^{-1} \sigma_x^2 I_{n_x \times n_x}\Big)\,,\\
	\pi(x | \sigma_x^2) &= \Ncal(x | 0, (\sigma_x^2 + \lambda_\zeta^2)  I_{n_x \times n_x})\,.
\end{aligned}
\end{align}

If we consider, for instance, images as the inputs to our neural network a natural choice for $\lambda_\zeta$ would be such that the bulk of $\pi(\zeta)$ covers the pixel range. For other cases there might be no natural choice or a priori knowledge available for $\lambda_\zeta$ and the noninformative limit $\lambda_\zeta \rightarrow \infty$, cf. Section \ref{subsec:variational_inference_for_lambda_zetarightarrow_infty}, can be more suitable.

\subsection{Variational inference for $\lambda_\zeta\rightarrow \infty$}
\label{subsec:variational_inference_for_lambda_zetarightarrow_infty}

 In this section we describe how to use the noninformative prior that arises from the setup in Section \ref{subsec:informative_distribution_for_pizeta} for $\lambda_\zeta \rightarrow \infty$. In this limit the marginal $\pi(x_i | \sigma_x^2)$ as in \eqref{eq:posterior_zeta} is obviously improper. The following lemma shows that this is not really problem and summarizes the actual algorithm used in Section \ref{sec:experiments} of this work.

\begin{lemma}
	Consider the setup from Section \ref{sec:an_eiv_model_for_deep_learning} and \ref{subsec:informative_distribution_for_pizeta}. In the limit $\lambda_\zeta\rightarrow \infty$ the distributions $\pi(\zeta|x, \sigma_x^2)$ and the posterior $\pi(\theta|\mathcal{D}, \ssigma^2)$ both converge\footnote{Mathematically speaking we here mean pointwise convergence of densities, which implies, by Scheff\'e's Lemma, convergence in distribution.} to a meaningful limit, namely
	\begin{align*}
		\pi(\zeta |x,\sigma_x^2) &= \Ncal\Big( \zeta_i \Big| x_i, \sigma_x^2 I_{n_x \times n_x}\Big)\,,\\
		\pi(\theta|\mathcal{D}, \ssigma^2) &= \frac{\pi(\theta) \rho(\theta; \mathcal{D})}{\int \dd \theta \, \pi(\theta) \rho(\theta;\mathcal{D})}\,.
	\end{align*}
	where $\rho(\theta; \mathcal{D} ) = \prod_{i=1}^N\int \dd \zeta_i\, \pi(\zeta_i | x_i, \sigma_x^2) p(y_i|\zeta_i,\theta, \sigma_y^2)$. For a variational distribution $q_\phi(\theta)$ the Kullback-Leibler divergence $\KL(q_\phi(\theta)|\pi(\theta|\mathcal{D},\sigma^2))$ coincides, up to terms independent of $\phi$, with the expression \eqref{eq:loss}.
\end{lemma}
\begin{proof}
	For $\pi(\zeta|x,\sigma_x^2)$ this directly follows from \eqref{eq:posterior_zeta}. Concerning the posterior $\pi(\theta|\mathcal{D}, \ssigma^2) $ note that with \eqref{eq:eiv_marginal_likelihood} we have
\begin{align*}
	\pi(\DD | \theta,\ssigma^2) &=\prod_{i=1}^N  \pi(x_i|\sigma_x^2) \int \dd \zeta_i\, \pi(\zeta_i | x_i, \sigma_x^2) p(y_i|\zeta_i,\theta, \sigma_y^2)  \\
				    &=\prod_{i=1}^N  \pi(x_i|\sigma_x^2) \,\cdot \, \prod_{i=1}^N \int \dd \zeta_i\, \pi(\zeta_i | x_i, \sigma_x^2) p(y_i|\zeta_i,\theta, \sigma_y^2) = c \,  \cdot \rho(\theta;\mathcal{D}) \,,
\end{align*}
where we wrote $c=\prod_{i=1}^N  \pi(x_i|\sigma_x^2)$. Plugging this into \eqref{eq:Bayes} we obtain indeed
\begin{align*}
	\pi(\theta|\DD, \ssigma^2) = \frac{\pi(\theta) \pi(\DD | \theta,\ssigma^2) }{\int \pi(\theta) \pi(\DD | \theta,\ssigma^2)} = \frac{c\cdot \pi(\theta)\rho(\theta;\mathcal{D})}{c \cdot\int \dd \theta \,\pi(\theta)\rho(\theta;\mathcal{D})} = 
	\frac{\pi(\theta)\rho(\theta;\mathcal{D})}{\int \dd \theta \,\pi(\theta)\rho(\theta;\mathcal{D})}
\,,
\end{align*} 
which is well-defined in the limit $\lambda_\zeta \rightarrow \infty$.
Using this expression one now easily reshapes $\KL(q_\phi(\theta)|\pi(\theta|\mathcal{D},\sigma^2))$, similar as in Section \ref{subsec:variational_inference_for_informative_pizeta}, to $\mathcal{L}(\phi) +  \log \int \dd \theta \,\pi(\theta)\rho(\theta;\mathcal{D})$. 
\end{proof}

\section{Details on training data}
\label{sec:details_on_training_data}
\subsection{Simulated data}
\label{subsec:simulated_data}
The following table lists the simulated data sets used for the experiments in Section \ref{sec:experiments}. They were constructed, in analogy to \eqref{eq:new_model}, via a ground truth function $g$ and
\begin{align}
	x = \zeta + \eps_x,\, y = g(\zeta) + \eps_y
\end{align}
where $\eps_x\sim \Ncal(0,\sigma_x^2 I_{n_x \times n_x}), \,\eps_y \sim \Ncal(0,\sigma_y^2 I_{n_y \times n_y})$ and where the $\zeta$ are uniformly drawn from a certain range, cf. Table \ref{tab:sim_data} below.
\hspace{0.5cm}
\begin{table}[h]
\begin{tabular}{l||ccccc}
	data set & $g(\zeta)$ & $\zeta$ range & $\sigma_x$ & $\sigma_y$ & data points    \\
	\hline
	linear  & $\zeta$ & $[-1,1]$ & $0.1$ & $0.2$ & 500   \\
	quadratic  & $\zeta^2$ & $[-1,1]$ & $0.1$ & $0.1$ & 500    \\
	cubic  & $\zeta^3$ & $[-1,1]$ & $0.2$  & $0.05$ & 1000   \\
	sine  & $\zeta+\sin(2\pi\zeta)+\sin(4\pi\zeta)$ & $[-0.2,0.8]$ & $0.04$ & $0.01$ & 2000 
\end{tabular}
\caption{Details on the simulated data sets used in this work.}
\label{tab:sim_data}
\end{table}
\hspace{0.5cm}

As explained in Section \ref{sec:experiments} we performed 10 different training runs for each method (EiV and non-EiV) and each data set, for which the data above was 10 times randomly split into a training set ($80\%$ of the data points) and a test set ($20\%$ of the data points). 

In figure \ref{subfig:true_cov_bar} we also presented results of a Bayesian regression based on functions similar to the $g$ listed in Table \ref{tab:sim_data}. In detail, we used $\tilde{g}(x)=\sum_{i=0}^K a_i x^i$ with $K=1$(for linear), $K=2$\,(for quadratic), $K=3$\,(for cubic) and
$\tilde{g}(x)= a_0 x+a_1 \sin(2\pi x)+a_2 \sin(4\pi x)$ (for sine). From $\tilde{g}$ we constructed the statistical model $p(y|x,a_1,a_2,\ldots)=\Ncal(\tilde{g}(x), \sigma_y^2)$, with the according $\sigma_y$ from table \ref{tab:sim_data}, and used it together with the noninformative prior $\pi(a_1,a_2,\ldots)\propto 1$ to obtain the posterior predictive distribution for $y$ given an input $x$ and the training data. Estimate and uncertainty are then taken to be the mean and standard deviation of the posterior predictive distribution.

\subsection{Real data}
\label{subsec:real_data}

The real data sets used in this work are all freely available and rather standard in the literature about uncertainty quantification in deep learning. We refer to \cite{hernandez2015probabilistic,lakshminarayanan2016simple,gal2016dropout}. Similarly to the simulated data from Section \ref{subsec:simulated_data}, we split for each of the 10 training runs the data randomly into a training set ($80\%$ of the data points) and a test set ($20\%$ of the data points).
In contrast to \cite{hernandez2015probabilistic,lakshminarayanan2016simple,gal2016dropout} we used the California housing data set \cite{pace1997sparse} instead of the Boston housing data set.

\section{Network and training details}
\label{sec:network_and_training_details}

\begin{figure}[h]
	\centering
	\includegraphics[width=0.4\textwidth]{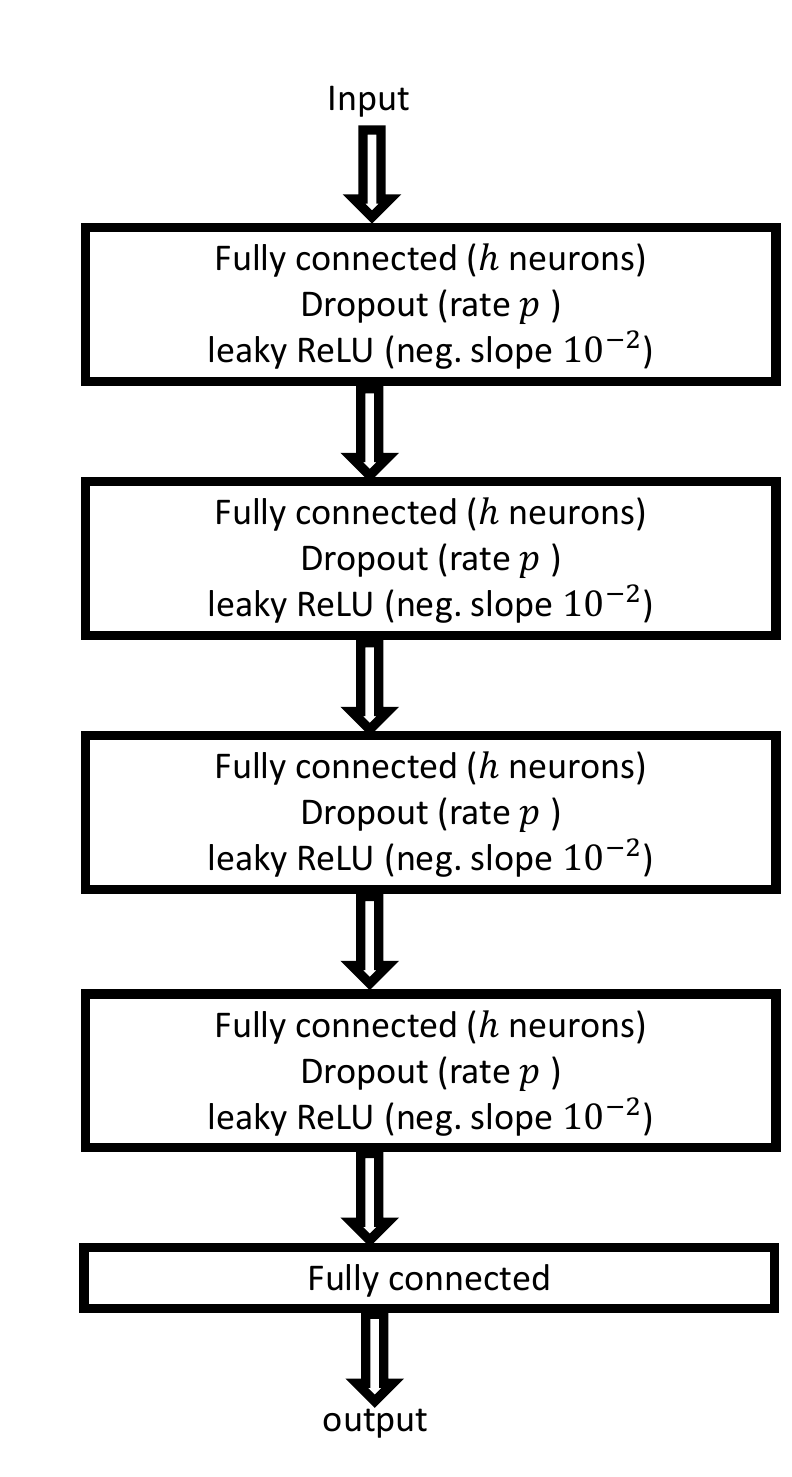}
	\caption{Sketch of the architecture of the networks used in this work. The precise architecture depends on two hyperparameters: the number of hidden units $h$ and the dropout rate $p$. Both are specified in Table \ref{tab:network_and_training}.}
	\label{fig:architecture}
\end{figure}

Throughout this work we used fully connected neural networks of the architecture sketched in Figure \ref{fig:architecture}. The precise architecture depends on two hyperparameters: the number of neurons $h$ in a hidden layer and the dropout rate $p$. Both are specified, together with details on the training, in Table \ref{tab:network_and_training} below. For all data sets and both approaches, EiV and non-EiV, we used an Adam optimizer and a $L^2$ regularization of $\frac{1}{2\lambda_\theta^2}=10$, with the notation of Section \ref{subsec:choosing_pitheta_pizeta_i_and_q_phi}. 

For both, the EiV and the non-EiV model, we used $M=1$ parameter samples in each training step and $M=100$ samples during evaluation. For the EiV model we used $L=5$ samples of $\zeta$ for both, evaluation and training (cf. Section \ref{sec:an_eiv_model_for_deep_learning}).

\begin{table}[h]
\begin{tabular}{l||cc|cccccc}
	data set & $h$ & $p$ & l.r. & batch & epochs & $\sigma_x$ & init. $\sigma_y$ & $n_{\mathrm{update}\,\sigma_y}$  \\
	\hline
		yacht & 1024 & 0.2 & $10^{-3}$ & 32 & 1200 & 0.05 & 0.5 & 500 \\
		naval & 1024 & 0.2 & $10^{-3}$ & 32 & 30 & 0.025 & 0.5 & 14 \\
		linear & 128 & 0.1 & $10^{-3}$ & 16 & 100 & 0.1 & 0.1 & 40 \\
		protein & 1024 & 0.2 & $10^{-3}$ & 100 & 30 & 0.05 & 0.5 & 14 \\
		concrete & 1024 & 0.2 & $10^{-3}$ & 32 & 100 & 0.05 & 0.5 & 40 \\
		kin8nm & 1024 & 0.2 & $10^{-3}$ & 32 & 30 & 0.05 & 0.5 & 14 \\
		wine & 1024 & 0.2 & $10^{-3}$ & 32 & 100 & 0.05 &  0.5 & 40 \\
		quadratic & 128 & 0.1 & $10^{-3}$ & 16 & 100 & 0.1 & 0.1 & 40 \\
		power & 1024 & 0.2 & $10^{-3}$ & 64 & 35 & 0.05 & 0.5 & 15 \\
		sine & 128 & 0.1 & $10^{-3}$ & 16 & 100 & 0.04 & 0.01 & 40 \\
		energy & 1024 & 0.2 & $10^{-3}$ & 32 & 600 & 0.05 & 0.5 & 250 \\
		california & 1024 & 0.1 & $10^{-3}$ & 200 & 100 & 0.05 & 0.5 & 40 \\
		cubic & 128 & 0.1 & $10^{-3}$ & 16 & 100 & 0.2 & 0.05 & 40 \\
\end{tabular}
\caption{Details on the network architecture, cf. Fig. \ref{fig:architecture}, and on the training: $h$ is the number of neurons in a hidden layer and $p$ the dropout rate (cf. Figure \ref{fig:architecture}), \emph{l.r.} is the learning rate, \emph{batch} the batch size, \emph{epochs} the epoch number, $\sigma_x$ the according parameter for the EiV approach, \emph{init.} $\sigma_y$ the used value for $\sigma_y$ prior to updating and $n_{\mathrm{update}\,\sigma_y}$ the number of epochs between two $\sigma_y$ updates (cf. Algorithm \ref{alg:train}).}
\label{tab:network_and_training}
\end{table}

\end{document}